\documentclass[runningheads]{llncs}
\usepackage{graphicx}
\usepackage{amsmath,amssymb} % define this before the line numbering.
\usepackage{color}
\usepackage[width=122mm,left=12mm,paperwidth=146mm,height=193mm,top=12mm,paperheight=217mm]{geometry}
\usepackage{algpseudocode} %allow you to write procedure and function pseudo codes
\usepackage{amsmath}
\usepackage{enumerate}
\usepackage{verbatim}
\usepackage{array} % make better tables
\usepackage{booktabs} % make better tables
\usepackage{xspace}

\newcommand{\paracompact}{\noindent\textbf}

\DeclareRobustCommand{\etal}{\textit{et al}.\@\xspace}
\DeclareRobustCommand{\ie}{\textit{i}.\textit{e}.\@\xspace}
\DeclareRobustCommand{\eg}{\textit{e}.\textit{g}.\@\xspace}

\newcommand{\xiao}{\textcolor{black}}
\newcommand{\xiaoa}{\textcolor{black}} %Add citation to Ali and Berg's work.
\newcommand{\cameraready}{\textcolor{black}}

\begin{document}
% \linenumbers
\pagestyle{headings}
\mainmatter

\title{Leveraging Visual Question Answering for Image-Caption Ranking} % Replace with your title

%\titlerunning{ECCV-16 submission ID \ECCV16SubNumber}

\authorrunning{Xiao Lin and Devi Parikh}

\author{Xiao Lin \quad Devi Parikh}
\institute{Bradley Department of Electrical and Computer Engineering,\\
Virginia Tech\\
\email{ \{linxiao,parikh\}@vt.edu}}

\maketitle

\begin{abstract}
%Visual Question Answering (VQA) is an ``AI-complete'' task that queries knowledge about multiple disciplines.
%Today's VQA models, while far from perfect, may already be picking up on such knowledge and may serve as an implicit knowledge resource to help other tasks.
%On the other hand image-caption ranking requires rich commonsense knowledge but inferring such knowledge from image-caption pairs is still challenging.
%In this work we propose to leverage knowledge in VQA corpora for image-caption ranking by using VQA models to assess probabilities of question-answer pairs on images and captions as features. These semantically meaningful VQA-grounded features interpret images and captions from many different perspectives and imagine beyond classical perception to better understand images and captions. 
%We propose score-level fusion and representation-level fusion approaches to incorporate VQA knowledge in VQA-agnostic image-caption ranking models using the VQA-grounded features.
%Our proposed approaches significantly outperform state-of-the-art VQA-agnostic models and set new state-of-the-arts on MSCOCO image-caption ranking by large margins.

Visual Question Answering (VQA) is the task of taking as input an image and a free-form natural language question about the image, and producing an accurate answer. In this work we view VQA as a ``feature extraction'' module to extract image and caption representations. We employ these representations for the task of image-caption ranking. Each feature dimension captures (imagines) whether a fact (question-answer pair) could plausibly be true for the image and caption. This allows the model to interpret images and captions from a wide variety of perspectives. We propose score-level and representation-level fusion models to incorporate VQA knowledge in an existing state-of-the-art VQA-agnostic image-caption ranking model. We find that incorporating and reasoning about consistency between images and captions significantly improves performance. Concretely, our model improves state-of-the-art on caption retrieval by 7.1\% and on image retrieval by 4.4\% on the MSCOCO dataset.

\end{abstract}

%%%%%%%%% BODY TEXT

\section{Introduction}
\label{sec:intro}

Visual Question Answering (VQA) is an ``AI-complete'' problem that requires knowledge from multiple disciplines such as computer vision, natural language processing and knowledge base reasoning. 
A VQA system takes as input an image and a free-form open-ended question about the image and outputs the natural language answer to the question.
A VQA system needs to not only recognize objects and scenes but also reason beyond low-level recognition about aspects such as intention, future, physics, material and commonsense knowledge. 
For example ($Q$: Who is the person in charge in this picture? $A$: Chef) reveals the most important person and occupation in the image. 
Moreover, answers to multiple questions about the same image can be correlated and may reveal more complex interactions. For example ($Q$: What is this person riding? $A$: Motorcycle) and ($Q$: What is the man wearing on his head? $A$: Helmet) might reveal correlations observable in the visual world due to safety regulations.

\begin{figure}
\centering
   \includegraphics[width=1.0\linewidth]{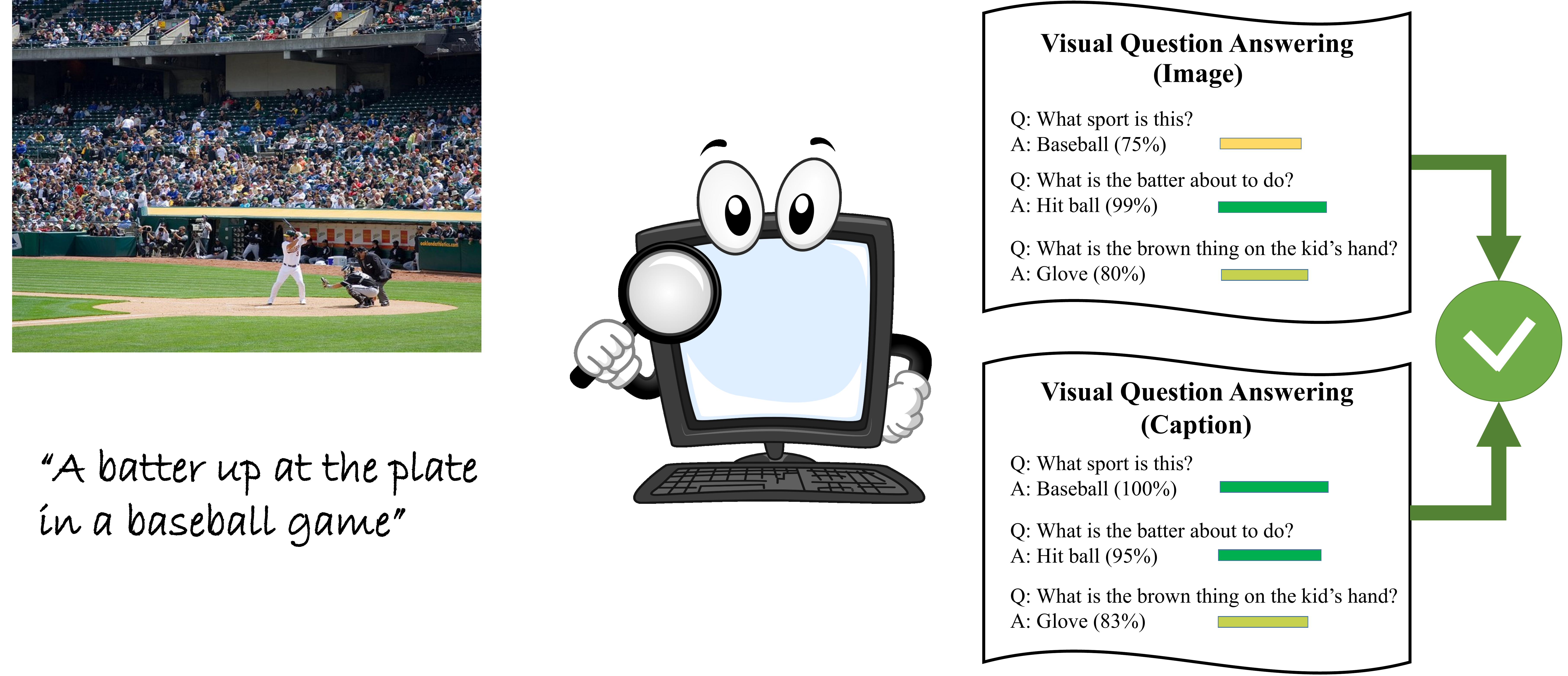}
   \caption{Aligning images and captions requires high-level reasoning \eg ``a batter up at the plate'' would imply that a player is holding a bat, posing to hit the baseball and there might be another player nearby waiting to catch the ball. There is rich knowledge in Visual Question Answering (VQA) corpora containing human-provided answers to a variety of questions one could ask about images. We propose to leverage knowledge in VQA by using VQA models learned on images and captions as ``feature extraction'' modules for image-caption ranking. }
\label{fig:teaser}
\end{figure}

Today's VQA models, while far from perfect, may already be picking up on these semantic correlations of the world. 
If so, they may serve as an implicit knowledge resource to help other tasks. 
Just like we do not need to fully understand the theory behind an equation to use it, can we already use VQA knowledge captured by existing VQA models to improve other tasks?

In this work we study the problem of using VQA knowledge to improve image-caption ranking. Consider the image and its caption in Figure~\ref{fig:teaser}. Aligning them not only requires recognizing the batter and that it is a baseball game (mentioned in the caption), but also realizing that a batter up at the plate would imply that a player is holding a bat, posing to hit the baseball and there might be another player nearby waiting to catch the ball (seen in the image). Image captions tend to be generic. As a result, image captioning corpora may not capture sufficient details for models to infer this knowledge.

%Inferring such knowledge from image-caption training data is still a challenging problem even with recent advances in deep learning. 

Fortunately VQA models try to explicitly learn such knowledge from a corpus of images, each with associated questions and answers. 
Questions about images tend to be much more specific and detailed than captions.
The VQA dataset of~\cite{Antol_2015} in particular has a collection of free-form open-ended questions and answers provided by humans.
These images also have associated captions~\cite{TYLin_2014}.

We propose to leverage VQA knowledge captured by such corpora for image-caption ranking by using VQA models learned on images and captions as ``feature extraction'' schemes to represent images and captions. Given an image and a caption, we choose a set of free-form open-ended questions and use VQA models learned on images and captions to assess probabilities of their answers. We use these probabilities as image and caption features respectively. In other words, we embed images and captions into the space of VQA questions and answers using VQA models. Such VQA-grounded representations interpret images and captions from a variety of different perspectives and imagine beyond low-level recognition to better understand images and captions.

We propose two approaches that incorporate these VQA-grounded representations into an existing 
%to use these VQA-grounded representations to add VQA knowledge to an existing 
state-of-the-art\footnote{To the best of our knowledge on MSCOCO~\cite{TYLin_2014}, \cite{Kiros_2015} has the state-of-the-art caption retrieval performance. \cite{Ma_2015} has the state-of-the-art image retrieval performance. } VQA-agnostic image-caption ranking model~\cite{Kiros_2015}: fusing their predictions and fusing their representations. We show that such VQA-aware models significantly outperform the VQA-agnostic model and set state-of-the-art performance on MSCOCO image-caption ranking. Specifically, we improve caption retrieval by 7.1\% and image retrieval by 4.4\%.

%Our results also demonstrate that knowledge in VQA corpora can be used as effectively as captions for image-caption ranking.

%%%%%%%%%%%%%%%%%%%%%%%%%%%%%%%%%%%%%%%%%%%%%%%%%%%%%%%%%%%%%%%%%%%%%
This paper is organized as follows: 
Section~\ref{sec:related_work} introduces related works. We first introduce VQA and image-caption ranking tasks as our building blocks in Section~\ref{sec:background}, then detail our VQA-based image-caption ranking models in Section~\ref{sec:approach}. Experiments and results are reported in Section~\ref{sec:result}. We conclude in Section~\ref{sec:discussion}.
\section{Related Work}
\label{sec:related_work}

\paracompact{Visual Question Answering.}
Visual Question Answering (VQA)~\cite{Antol_2015} is the task of taking an image and a free-form open-ended question about the image and automatically predicting the natural language answer to the question. 
VQA may require fine-grained recognition, object detection, activity recognition, multi-modal and commonsense knowledge. Large datasets~\cite{Malinowski_2014,Ren_2015,Yu_2015,Gao_2015,Antol_2015} have been made available to cover the diversity of knowledge required for VQA. Most notably the VQA dataset~\cite{Antol_2015} contains 614,163 questions and ground truth answers on 204,721 images of the MSCOCO~\cite{TYLin_2014} dataset.

Recent VQA models~\cite{Malinowski_2015,Ren_2015,Gao_2015,Zhou_2015,Antol_2015,Ma_2015} explore state-of-the-art deep learning techniques combining Convolutional Neural Networks (CNNs) and Recurrent Neural Networks (RNNs). 
~\cite{Antol_2015} also explores a slight variant of VQA that answers a question about the image by reading a caption describing the image instead of looking at the image itself. We call this variant VQA-Caption.
%\cameraready{Interestingly, VQA models using captions instead of images perform better than models that use images, perhaps because current vision systems are unable to understand images accurately enough, but AI systems are capable of understanding single-sentence captions reasonably well. }

VQA is a challenging task in its early stages. In this work we propose to use both VQA and VQA-Caption models as implicit knowledge resources. We show that current VQA models, while far from perfect, can already be used to improve other multi-modal AI tasks; specifically image-caption ranking. 

\paracompact{Semantic mid-level visual representations.}
Previous works have explored the use of attributes~\cite{Farhadi_2009,Branson_2010,Wang_2010}, parts~\cite{Berg_2013,Zhang_2013_ICCV}, poselets~\cite{Bourdev_2010,Zhang_2014}, objects~\cite{Li_2010}, actions~\cite{Sadanand_2012} and contextual information~\cite{Gupta_2008,Tang_2015,Doersch_2015} as sematic mid-level representations for visual recognition.
Benefits of using such semantic mid-level visual representations include improving fine-grained visual recognition, learning models of visual concepts without example images (zero-shot learning~\cite{Lampert_2009,Parikh_2011}) and improving human-machine communication where a user can explain the target concept during image search~\cite{Kumar_2011,Kovashka_2012}, or give a classifier an explanation of labels~\cite{Donahue_2011,Parkash_2012}.
Recent works also explore using word embeddings~\cite{Socher_2013} and free-form text~\cite{Elhoseiny_2013} as representations for zero-shot learning of new object categories. \cite{Johnson_2015} proposes scene graphs for image retrieval. \cite{Antol_2014} proposes using abstract scenes as an intermediate representation for zero-shot action recognition. 
\xiaoa{Closest to our work is the use of objects, actions, scenes~\cite{Farhadi_2010}, attributes and object interactions~\cite{Kulkarni_2011} for generating and ranking image captions.}
In this work we propose to use free-form open-ended questions and answers as mid-level representations and we show that they provide rich interpretations of images and captions. 

\paracompact{Commonsense knowledge for visual reasoning.}
Recently there has been a surge of interest in visual reasoning tasks that require high-level reasoning such as physical reasoning~\cite{Hamrick_2011,Zheng_2013}, future prediction~\cite{Fouhey_2014,Walker_2014,Pirsiavash_2014}, object affordance prediction~\cite{Zhu_2014} and textual tasks that require visual knowledge~\cite{Lin_2015,Vedantam_2015_ICCV,Sadeghi_2015}. 
Such tasks can often benefit from reasoning with external commonsense knowledge resources.
\cite{Zhu_2015} uses a knowledge base learned on object categories, \xiao{attributes}, actions and \xiao{object affordances} for query-based image retrieval.
\cite{Vondrick_2015} learns to anticipate future scenes from watching videos for action and object forecasting. 
\cite{Lin_2015} learns to imagine abstract scenes from text for textual tasks that need visual understanding.
\cite{Vedantam_2015_ICCV,Sadeghi_2015} evaluate the plausibility of commonsense assertions by verifying them on collections of abstract scenes and real images, respectively, to leverage the visual common sense in those collections.
Our work explores the use of VQA corpora which have both visual (image) and textual (captions) commonsense knowledge for image-caption ranking.

%Common sense is an important element in solving many beyond recognition tasks, since beyond recognition tasks tend to require information that is outside the boundaries of the image. It has been shown that learning and using non-visual common sense (i.e. common sense learnt from non-visual sources) benefits physical reasoning [23, 49], reasoning about intentions [40] and object functionality [50]. One instantiation of visual common sense that has been leveraged in the vision community in the past is the use of contextual reasoning for improved recognition [22, 12, 21, 17, 25, 50]. In this work, we explore the use of visual common sense for seemingly nonvisual tasks through “imagination”, i.e. generating scenes.

\paracompact{Images and captions.}
Recent works~\cite{Karpathy_2015,Chen_2015_CVPR,Kiros_2015,Xu_2015,Mao_2015,Ma_2015_ICCV} have made significant progress on automatic image caption generation and ranking by applying deep learning techniques for image recognition~\cite{Krizhevsky_2012,Simonyan_2014,Szegedy_2015} and language modeling~\cite{Cho_2014,Sutskever_2014} on large datasets~\cite{Deng_2009,TYLin_2014}. 
Algorithms can now often generate accurate, human-like natural-language captions for images. 
However, evaluating the quality of such automatically generated open-ended image captions is still an open research problem~\cite{Elliott_2014,Vedantam_2015_CVPR}. 

On the other hand, ranking images given captions and ranking captions given images require a similar level of image and language understanding, but are amenable to automatic evaluation metrics.
Recent works on image-caption ranking mainly focus on improving model architectures. 
\cite{Kiros_2015,Mao_2015} study different architectures for projecting CNN image representations and RNN caption representations into a common multi-modal space. \cite{Ma_2015_ICCV} uses multi-modal CNNs for image-caption ranking. \cite{Karpathy_2015} aligns image and caption fragments using CNNs and RNNs.
Our work takes an orthogonal approach to previous works. 
We propose to leverage knowledge in VQA corpora containing questions about images and associated answers for image-caption ranking. 
Our proposed VQA-based image and caption representations provide complementary information to those learned using previous approaches on a large image-caption ranking dataset.

\section{Building Blocks: Image-Caption Ranking and VQA}
\label{sec:background}

In this section we present image-caption ranking and VQA modules that we build on top of.

\subsection{Image-caption ranking}

The image-caption ranking task is to retrieve relevant images given a query caption, and relevant captions given a query image. During training we are given image-caption pairs $(I,C)$ that each corresponds to an image $I$ and its caption $C$. 
For each pair we sample $K-1$ other images in addition to $I$ so the image retrieval task becomes retrieving $I$ from $K$ images $I_i, i=1,2\ldots K $ given caption $C$. We also sample $K-1$ random captions in addition to $C$ so the caption retrieval task becomes retrieving $C$ from $K$ captions $C_i, i=1,2\ldots K $ given image $I$. 

Our image-caption ranking models learn a~\cameraready{ranking} scoring function $S(I,C)$ such that the corresponding retrieval probabilities:
\begin{equation}
  \label{eq:t}
  \begin{aligned}
    P_{im}(I|C)=\frac{\exp(S(I,C))}{\sum\limits_{i=1}^K \exp(S(I_{i},C))}
	\qquad
    P_{cap}(C|I)=\frac{\exp(S(I,C))}{\sum\limits_{i=1}^K \exp(S(I,C_{i}))}
  \end{aligned}
\end{equation}
are maximized. 
%Let $\theta$ denote the parameters to learn in $S(I,C)$. 
Let $S(I,C)$ be parameterized by $\theta$ (to be learnt). 
We formulate an objective function $L(\theta)$ for $S(I,C)$ as the sum of expected negative log-likelihoods of image and caption retrieval over all image-caption pairs $(I,C)$:
\begin{equation}
  \label{eq:loss}
  \begin{aligned}
    L(\theta)=\mathbb{E}_{(I,C)}[-\log P_{im}(I|C)] + \mathbb{E}_{(I,C)}[-\log P_{cap}(C|I)]
  \end{aligned}
\end{equation}
%At test time we rank images and captions using $S(I,C)$. 

Recent works on image-caption ranking often construct $S(I,C)$ by combining a vectorized image representation which is usually hidden layer activations in a CNN pretrained for image classification, with a vectorized caption representation which is usually a sentence encoding computed using an RNN in a multi-modal space. Such scoring functions rely on large image-caption ranking datasets to learn knowledge necessary for image-caption ranking and do not leverage knowledge in VQA corpora. We call such models VQA-agnostic models. 

In this work we use the publicly available state-of-the-art image-caption ranking model of~\cite{Kiros_2015} as our baseline VQA-agnostic model. \cite{Kiros_2015} projects a $D_{x_I}$-dimensional CNN activation $x_I$ for image $I$ and a $D_{x_C}$-dimensional RNN latent encoding $x_C$ for caption $C$ to the same $D_{x_C}$-dimensional common multi-modal embedding space as unit-norm vectors $t_I$ and $t_C$:

\begin{equation}
  \label{eq:t}
  \begin{aligned}
    t_I = \frac{W_{I} x_I}{||W_{I} x_I||_2} \qquad
    t_C = \frac{x_C}{||x_C||_2}
  \end{aligned}
\end{equation}
%\xiao{Here $W_{I}$ is a $D_t\times D_{x_I}$ matrix and $W_{C}$ is a $D_t\times D_{x_C}$ matrix.}
The multi-modal scoring function is defined as their dot product $S_t(I,C)=\langle t_I,t_C\rangle$.

The VQA-agnostic model of~\cite{Kiros_2015} uses the 19-layer VGGNet~\cite{Simonyan_2014} ($D_{x_I}=4096$) for image encoding and an RNN with $1024$ Gated Recurrent Units~\cite{Cho_2014} ($D_{x_C}=1024$) for caption encoding. The RNN and \xiao{parameters $W_{I}$} are jointly learned on the image-caption ranking training set using a margin-based objective function. 

%\subsection{Visual Question Answering}
\subsection{VQA}
\label{sec:vqa_model}

VQA is the task of given an image $I$ and a free-form open-ended question $Q$ about $I$, generating a natural language answer $A$ to that question. 
Similarly, VQA-Caption task proposed by \cite{Antol_2015} takes a caption $C$ of an image and a question $Q$ about the image, then generates an answer $A$.
In~\cite{Antol_2015} the generated answers are evaluated using $\min (\frac{\text{\# humans that provided }A}{3},1)$. That is, $A$ is 100\% correct if at least 3 humans (out of 10) provide the answer $A$. 

We closely follow~\cite{Antol_2015} and formulate VQA as a classification task over top $M=1000$ most frequent answers from the training set. 
The oracle accuracies of picking the best answer for each question within this set of answers are 89.37\% on training and 88.83\% on validation.
%These answers cover 85.08\% of training and 84.15\% of validation questions on the VQA dataset. 
During training, given triplets of image $I$, question $Q$ and ground truth answer $A$, we optimize the negative log-likelihood (NLL) loss to maximize the probability of the ground truth answer $P_{I}(A|Q,I)$ given by the VQA model. Similarly given triplets of caption $C$, question $Q$ and ground truth answer $A$, we optimize the NLL loss to maximize the VQA-Caption model probability $P_{C}(A|Q,C)$. 

Following~\cite{Antol_2015}, for a VQA question $(I,Q)$ we first encode the input image $I$ using the 19-layer VGGNet~\cite{Simonyan_2014} as a 4,096-dimensional image encoding $x_I$, and encode the question $Q$ using a 2-layer RNN with 512 Long Short-Term Memory (LSTM) units~\cite{Hochreiter_1997} per layer as a 2,048-dimensional question encoding $x_Q$. We then project $x_I$ and $x_Q$ into a common 1,024-dimensional multi-modal space as $z_I$ and $z_Q$:
\begin{equation}
  \label{eq:t}
  \begin{aligned}
    z_I = Tanh( W_{I} x_I + b_I ) \qquad
    z_Q = Tanh( W_{Q} x_Q + b_Q )
  \end{aligned}
\end{equation}
%\xiao{Here $W_{I}$ is a $1024\times 4096$ matrix, $b_I$ is a $1024\times 1$ vector, $W_Q$ is a $1024\times 2048$ matrix and $b_Q$ is a $1024\times 1$ vector.}

As in~\cite{Antol_2015} we then compute the representation $z_{I+Q}$ for the image-question pair $(I,Q)$ by element-wise multiplying $z_I$ and $z_Q$: $z_{I+Q}=z_I\odot z_Q$. The scores $s_A$ for 1,000 answers are given by:

\begin{equation}
  \label{eq:t}
  \begin{aligned}
	s_A = W_{s} z_{I+Q} + b_s
  \end{aligned}
\end{equation}
%\xiao{Here $W_s$ is a $1000\times 1024$ matrix and $b_s$ is a $1000\times 1$ vector.}
We jointly learn the question encoding RNN and parameters $\{W_I,b_I,W_Q,b_Q,W_s,$ $b_s\}$ during training.

For the VQA-Caption task given caption $C$ and question $Q$, we use the same network architecture and learning procedure as above, but using the most frequent 1,000 words in training captions as the dictionary to construct a 1,000 dimensional bag-of-words encoding for caption $C$ as $x_C$ to replace the image feature $x_I$ and compute $z_C$, $z_{C+Q}$ respectively.

The VQA and VQA-Caption models are learned on the train split of the VQA dataset~\cite{Antol_2015} using 82,783 images, 413,915 captions and 248,349 questions. 
These models achieve VQA validation set accuracies of 54.42\% (VQA) and 56.28\% (VQA-Caption), respectively. Next, they are used as sub-modules in our image-caption ranking approach.
%\cameraready{Interestingly, VQA models using captions instead of images perform better than models that use images, perhaps because current vision systems are unable to understand images accurately enough, but AI systems are capable of understanding single-sentence captions reasonably well. }

\begin{figure}
\begin{center}
   \includegraphics[width=0.85\linewidth]{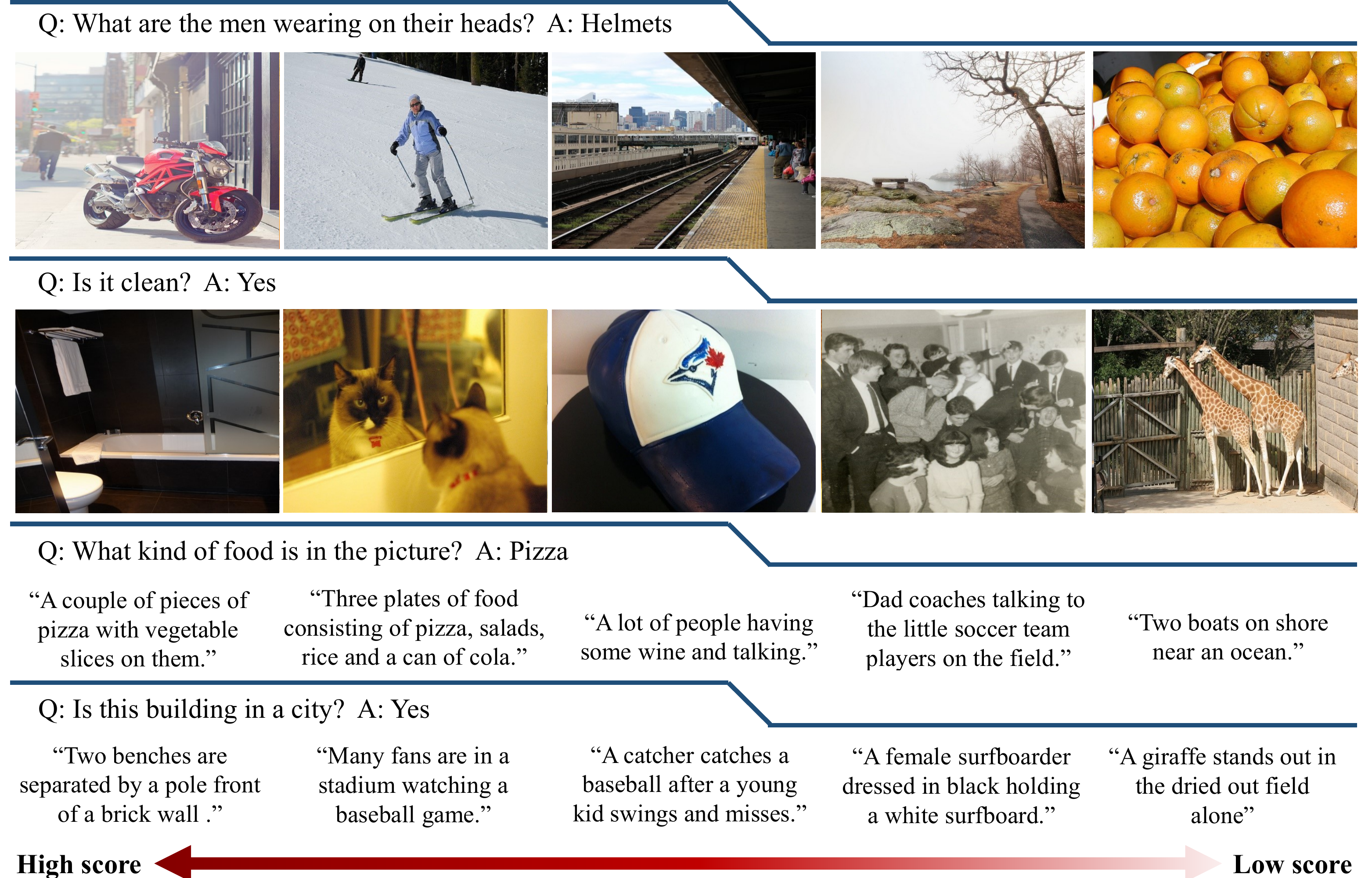}
\end{center}
   \caption{Images and captions sorted by $P_I(A|Q,I)$ and $P_C(A|Q,C)$ assessed by our VQA (top) and VQA-Caption (bottom) models respectively. Indeed, images and captions that are more plausible for the $(Q,A)$ pairs are scored higher.}
\label{fig:qual_vqa}
\end{figure}

\section{Approach}
\label{sec:approach}

To leverage knowledge in VQA for image-caption ranking, we propose to represent the images and the captions in the VQA space using VQA and VQA-Caption models. We call such representations VQA-grounded representations.

\subsection{VQA-grounded representations}
\label{sec:vqa_grounded_representations}

Let's say we have a VQA model $P_I(A|Q,I)$, a VQA-Caption model $P_C(A|Q,C)$ and a set of $N$ questions $Q_i$ and their plausible answers (one for each question) $A_i$, $i=1,2,...N$. Then given an image $I$ and a caption $C$, we first extract the $N$ dimensional VQA-grounded activation vectors $u_I$ for $I$ and $u_C$ for $C$ such that each dimension $i$ of $u_I$ and $u_C$ is the log probability of the ground truth answer $A_i$ given a question $Q_i$.

\begin{equation}
  \label{eq:t}
  \begin{aligned}
%    u_I^{(i)} &= \log P_I(A_i|Q_i,I), i=1,2,\ldots, N \\
    u_I^{(i)} = \log P_I(A_i|Q_i,I) \qquad
    u_C^{(i)} = \log P_C(A_i|Q_i,C), i=1,2,\ldots, N
  \end{aligned}
\end{equation}

For example if the $(Q_i,A_i)$ pairs are ($Q_1$: What is the person riding?, $A_1$: Motorcycle) and ($Q_2$: What is the man wearing on his head?, $A_2$: Helmet), $u_I^{(1)}$ and $u_C^{(1)}$ verify if the person in image $I$ and caption $C$ respectively is riding a motorcycle. At the same time $u_I^{(2)}$ and $u_C^{(2)}$ verify whether the man in $I$ and $C$ is wearing a helmet. \cameraready{Figure~\ref{fig:teaser} shows another example.}

%\qa{In free-form open-ended VQA, an interesting phenomena is that the $(Q,A)$ pairs are sometimes about the context or subjective evaluations of the image. For example ($Q$: Is this an old photo? $A$: Yes) and ($Q$: What is the bird's distinguishing feature? $A$: Color). While they are not directly about what is present in the image, they could provide useful cues to match an image with its captions.}

In cases where there is not a man in the image or the caption, \ie the assumption of $Q_i$ is not met, $P_I(A_i|Q_i,I)$ and $P_C(A_i|Q_i,C)$ may still reflect if there \emph{were} a man or if the assumption of $Q_i$ \emph{were} fulfilled, could he be wearing a helmet. 
In other words, even if there is no person present in the image or mentioned in the caption, the model may still assess the plausibility of a man wearing a helmet or a motorcycle being present.
This imagination beyond what is depicted in the image or caption can be helpful in providing additional information when reasoning about the compatibility between an image and a caption. 
We show qualitative examples of this imagination or plausibility assessment for selected $(Q,A)$ pairs in Figure~\ref{fig:qual_vqa} where we sort images and captions based on $P_I(A|Q,I)$ and $P_C(A|Q,C)$. Indeed, scenes where the corresponding fact $(Q,A)$ (e.g., man is wearing a helmet) is more likely to be plausible are scored higher. 
\footnote{\cameraready{Nonetheless, checking if a question applies to the target image and caption is also desirable. Contemporary work~\cite{Ray_2016} has looked at modeling $P(Q|I)$, and can be incorporated in our approach as an additional feature.}}

Based on the activation vectors $u_I$ and $u_C$, we then compute the VQA-grounded vector representations $v_I$ and $v_C$ for $I$ and $C$ by projecting $u_I$ and $u_C$ to a $D_u$-dimensional vector embedding space:

\begin{equation}
  \label{eq:t}
  \begin{aligned}
    v_I = \sigma (W_{u_I} u_I + b_{v_I} ) \qquad
    v_C = \sigma (W_{u_C} u_C + b_{v_C} )
  \end{aligned}
\end{equation}
Here $\sigma$ is a non-linear activation function. %\xiao{$W_{u_I}$ and $W_{u_C}$ are $D_u\times N$ matrices. $b_{v_I}$ and $b_{v_C}$ are $D_u\times 1$ vectors. }

%\xiao{Here we're like using the very last prediction layer to generate a representation. Note that a common practice of extracting image representation from pretrained ImageNet classification networks is to use the top layers as image representations and finetune them for the target tasks. In our case however, we don't know if that's a good idea or not. I think it's probably not because you might not be able to utilize correlations of different QAs from the same image. I'll conduct experiments to verify that.}

By verifying question-answer pairs on image $I$ and caption $C$ and computing vector representations on top of them, the VQA-grounded representations $v_I$ and $v_C$ explicitly project image and caption into VQA space to utilize knowledge in the VQA corpora. 
However, that comes at a cost of losing information such as the sentence structure of the caption and image saliency. These information can also be important for image-caption ranking. 
As a result, We find VQA-grounded representations are most effective when they are combined with baseline VQA-agnostic models, so we propose two strategies for fusing VQA-grounded representations with baseline VQA-agnostic models: combining their prediction scores or score-level fusion (Figure~\ref{fig:models} left) and combining their representations or representation-level fusion (Figure~\ref{fig:models} right).

%Figure~\ref{fig:models} illustrates the network architectures of our score-level and representation-level fusion models for image-caption ranking. 
%We now describe their details in Section~\ref{sec:score_level_fusion} and Section~\ref{sec:representation_level_fusion}.

\subsection{Score-level fusion}
\label{sec:score_level_fusion}

A simple strategy to combine our VQA-grounded model with a VQA-agnostic image-ranking model is to combine them at the score level. Given image $I$ and caption $C$, we first compute the VQA-grounded score as the dot product between the VQA-grounded representations of image and caption $S_v(I,C)=\langle v_I,v_C\rangle$. We then combine it with the VQA-agnostic scoring function $S_{t}(I,C)$ to get the final scoring function $S(I,C)$:

\begin{figure}
\begin{center}
   \includegraphics[width=1\linewidth]{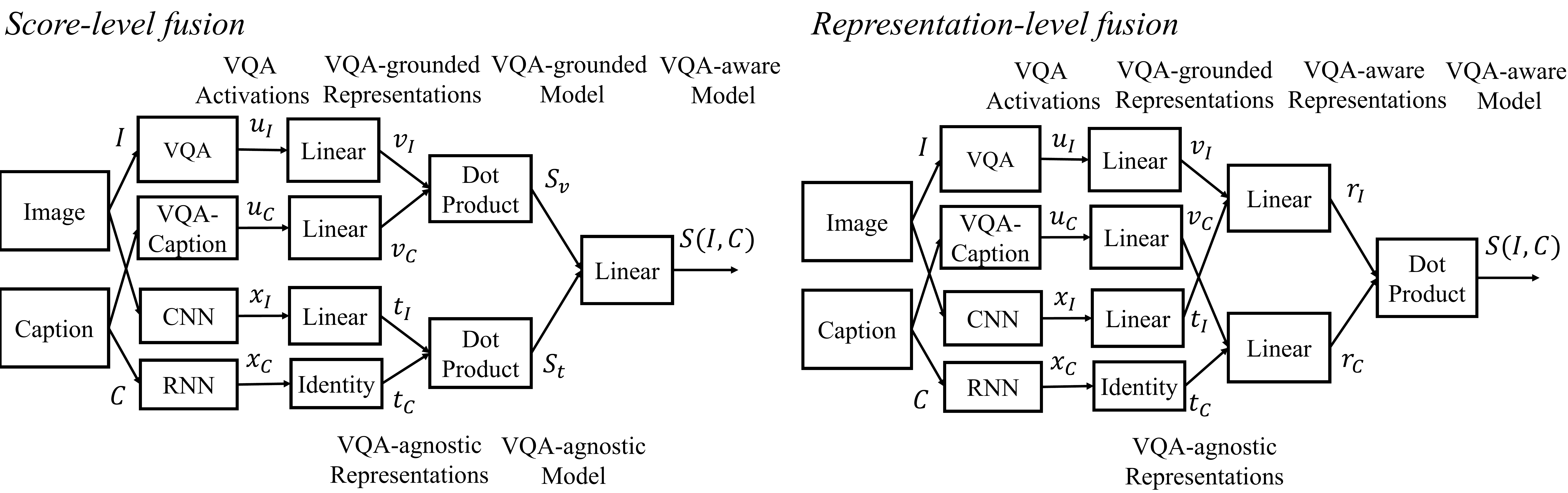}
\end{center}
   \caption{We propose score-level fusion (left) and representation-level fusion (right) to utilize VQA for image-caption ranking. They use VQA and VQA-Caption models as ``feature extraction'' schemes for images and captions and use those features to construct VQA-grounded representations. The score-level fusion approach combines the scoring functions of a VQA-grounded model and a baseline VQA-agnostic model. The representation-level fusion approach combines VQA-grounded representations and VQA-agnostic representations to produce a VQA-aware scoring function.}
\label{fig:models}
\end{figure}

\begin{equation}
  \label{eq:t}
  \begin{aligned}
    S(I,C)=\alpha S_{t}(I,C)+ \beta S_{v}(I,C)
  \end{aligned}
\end{equation}

We first learn $\{ W_{u_I}, b_{u_I}, W_{u_C}, b_{u_C}\}$ on the image-caption ranking training set, and then learn $\alpha$ and $\beta$ on a held out validation set to avoid overfitting.

\subsection{Representation-level fusion}
\label{sec:representation_level_fusion}

An alternative to combining the VQA-agnostic and VQA-grounded representations at the score level is to inject the VQA-grounding at the representation level. Given the VQA-agnostic $D_t$-dimensional image and caption representations $t_I$ and $t_C$ used by the baseline model, we first compute the VQA-grounded representations $v_I$ for image and $v_C$ for caption introduced in Section~\ref{sec:vqa_grounded_representations}. And then they are combined with VQA-agnostic representations to produce VQA-aware representations $r_I$ for image $I$ and $r_C$ for caption $C$ by projecting them to a $D_r$-dimensional multi-modal embedding space as follows:

\begin{equation}
  \label{eq:rep_fus}
  \begin{aligned}
    r_I = \sigma (W_{t_I} t_I + W_{v_I} v_I + b_{r_I} ) \qquad
    r_C = \sigma (W_{t_C} t_C + W_{v_C} v_C + b_{r_C} )
  \end{aligned}
\end{equation}
%\xiao{Here $W_{t_I}$ and $W_{t_C}$ are $D_r\times D_t$ matrices. $W_{v_I}$ and $W_{v_C}$ are $D_r\times D_v$ dimensional matrices. $b_{r_I}$ and $b_{r_C}$ are $D_r\times 1$ vectors.}

The final image-caption ranking score is then 
\begin{equation}
S(I,C)=\langle r_I, r_C\rangle
\end{equation}

In experiments, we jointly learn $\{ W_{u_I}, b_{u_I}, W_{u_C}, b_{u_C}\}$ (for projecting $u_I$ and $u_C$ to the VQA-grounded representations $v_I$, $v_C$) with $\{ W_{t_I}, W_{v_I}, b_{r_I},$ $W_{t_C}, W_{v_C},$ $b_{r_C}\}$ (for computing the combined VQA-aware representations $r_I$ and $r_C$) on the image-caption ranking training set by optimizing Eq.~\ref{eq:loss}.

Score-level fusion and representation-level fusion models are implemented as multi-layer neural networks. All activation functions $\sigma$ are $ReLU(x)=\max(x,0)$ (for speed) and dropout layers~\cite{Srivastava_2014} are inserted after all $ReLU$ layers to avoid overfitting. We set the dimensions of the multi-modal embedding spaces $D_v$ and $D_r$ to 4,096 so they are large enough to capture necessary concepts for image-caption ranking.
Optimization hyperparameters are selected on the validation set. We optimize both models using RMSProp with batch size 1,000 at learning rate 1e-5 for score-level fusion and 1e-4 for representation-level fusion. Optimization runs for 100,000 iterations with learning rate decay every 50,000 iterations.

Our main results in Section~\ref{sec:results_main} use $N=3000$ question-answer pairs, sampled 3 questions per image with their ground truth answers with respect to their original images from 1,000 random VQA training images. We discuss using different numbers of question-answer pairs $N$ and different strategies for selecting the question-answer pairs in Section~\ref{sec:results_qa_pairs}.

\section{Experiments and Results}
\label{sec:result}

We report results on MSCOCO~\cite{TYLin_2014} which is the largest available image-caption ranking dataset. Following the splits of~\cite{Karpathy_2015,Kiros_2015} we use all 82,783 MSCOCO train images with 5 captions per image as our train set, 413,915 image-caption pairs in total. Note that this is the same split as the train split in the VQA dataset~\cite{Antol_2015} we used to train our VQA and VQA-Caption models. The validation set consists of 1,000 images sampled from the original MSCOCO validation images. The test set consists of 5,000 images sampled from the original MSCOCO validation images that were not in the image-caption ranking validation set. Same as the train set, there are 5 captions available for each validation and test image.

We follow the evaluation metric of~\cite{Karpathy_2015} and report caption and image retrieval performances on the first 1,000 test images following \cite{Karpathy_2015,Klein_2015,Mao_2015,Ma_2015,Kiros_2015}. 
Given a test image, the caption retrieval task is to find any 1 out of its 5 captions from all 5,000 test captions. 
Given a test caption, the image retrieval task is to find its original image from all 1,000 test images. We report recall@(1, 5, 10): the fraction of times a correct item was found among the top (1, 5, 10) predictions.

\subsection{Image-caption ranking results}
\label{sec:results_main}

Table~\ref{table:qa_ranking_result} shows our main results on MSCOCO. 
%Powered by knowledge in the VQA corpora, 
Our score-level fusion VQA-aware model using $N=3000$ question-answer pairs (``$N=3000$ score-level fusion VQA-aware'') achieves 46.9\% caption retrieval recall@1 and 35.8\% image retrieval recall@1. This model shows an improvement of 3.5\% caption and 4.8\% image retrieval recall@1 over the state-of-the-art VQA-agnostic model of~\cite{Kiros_2015}.

\begin{table}
\centering
\caption{Caption retrieval and image retrieval performances of our models compared to baseline models on MSCOCO image-caption ranking test set. Powered by knowledge in VQA corpora, both our score-level fusion and representation-level fusion VQA-aware approaches outperform state-of-the-art VQA-agnostic models by a large margin}
\begin{tabular}{ l |c c c|c c c }
\toprule
\multicolumn{7}{c}{\textbf{MSCOCO}}\\
\midrule
\multicolumn{1}{c}{Approach}&  \multicolumn{3}{c}{Caption Retrieval} & \multicolumn{3}{c}{Image Retrieval}\\
    & R@1   & R@5    & R@10   & R@1 & R@5 & R@10 \\
\midrule
Random  &  0.1 & 0.5  & 1.0  & 0.1 & 0.5 & 1.0 \\
DVSA~\cite{Karpathy_2015}  &  38.4 & 69.9  & 80.5  & 27.4 & 60.2 & 74.8 \\
FV (GMM+HGLMM)~\cite{Klein_2015} & 39.4 & 67.9 & 80.9 & 25.1 & 59.8 & 76.6 \\
$m$-RNN-vgg~\cite{Mao_2015}   &  41.0 & 73.0  & 83.5  & 29.0 & 42.2 & 77.0 \\
$m$-$CNN_{ENS}$~\cite{Ma_2015}   &  42.8 & 73.1   & 84.1  & 32.6 & 68.6 & 82.8\\
%\upd{Order embedding (1-crop)~\cite{Vendrov_2016}}   &  \upd{41.4} & \upd{-}  & \upd{84.2}  & \upd{33.5} & \upd{-} & \upd{82.2}\\
Kiros \etal~\cite{Kiros_2015} (VQA-agnostic)   &  43.4 & 75.7  & 85.8  & 31.0 & 66.7 & 79.9\\
\midrule
N=3000 score-level fusion VQA-grounded only  & 37.0 &  67.9 & 79.4 &  26.2 & 60.1   & 74.3 \\
N=3000 score-level fusion VQA-aware &  46.9  &  78.6   & 88.9  & 35.8 & 70.3 & \textbf{83.6} \\
\midrule
N=0 representation-level fusion VQA-agnostic  &  45.8 & 76.8  & 86.1  & 33.6 & 67.8 & 81.0\\
N=3000 representation-level fusion VQA-aware  &  \textbf{50.5}  &  \textbf{80.1}   & \textbf{89.7}  & \textbf{37.0} & \textbf{70.9} & 82.9\\
					
\bottomrule
\end{tabular}
\label{table:qa_ranking_result}
\end{table}

Our representation-level fusion approach adds an additional layer on top of the VQA-agnostic representations, resulting in a deeper model, so we experiment with adding an additional layer to the VQA-agnostic model for a fair comparison. That is equivalent to representation-level fusion using $N=0$ question-answer pair (``$N=0$ representation-level fusion'', \ie deeper VQA-agnostic). Comparing with the VQA-agnostic model of ~\cite{Kiros_2015}, adding this additional layer improves performance by 2.4\% caption and 2.6\% image retrieval recall@1.

By leveraging VQA knowledge our ``$N=3000$ representation-level fusion VQA-aware'' model achieves 50.5\% caption retrieval recall@1 and 37.0\% image retrieval recall@1, which further improves 4.7\% and 3.4\% over the $N=0$ VQA-agnostic representation-level fusion model. These improvements are consistent with our score-level fusion approach so this shows that the VQA corpora consistently provide complementary information to image-caption ranking.

To the best of our knowledge, the $N=3000$ representation-level fusion VQA-aware result is the best result on MSCOCO image-caption ranking and significantly surpasses previous best results by as much as 7.1\% in caption retrieval and 4.4\% image retrieval recall@1.

Our VQA-grounded model alone (``$N=3000$ score-level fusion VQA-grounded only'') achieves 37.0\% caption and 26.2\% image retrieval recall@1. This indicates that the VQA activations $u_I$ and $u_C$ which evaluate the plausibility of facts (question-answer pairs) in images and captions are informative representations.

\begin{figure}
\begin{center}
   \includegraphics[width=0.83\linewidth]{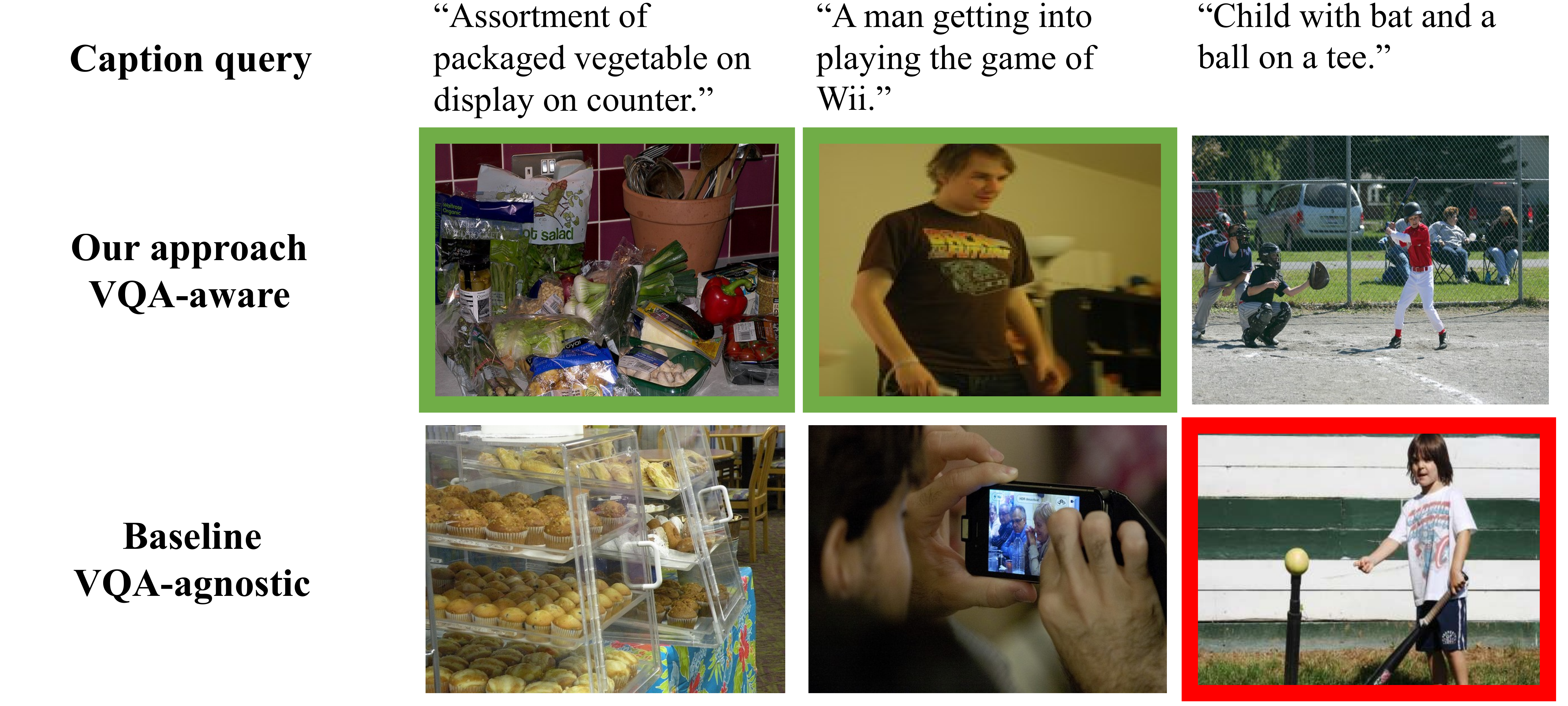}
\end{center}
   \caption{Qualitative image retrieval results of our score-level fusion VQA-aware model (middle) and the VQA-agnostic model (bottom). The true target image is highlighted (green if VQA-aware found it, red if VQA-agnostic found it but VQA-aware did not).}
\label{fig:qual}
\end{figure}

Figure~\ref{fig:qual} shows qualitative results on image retrieval comparing our approach ($N=3000$ score-level fusion) with the VQA-agnostic model. 
By looking at several top retrieved images from our model for the failure case (last column), we find that our model seems to have picked up on a correlation between bats and helmets. It seems to be looking for helmets in retrieved images, while the ground truth image does not have one.

We also experiment with using the hidden activations available in the VQA and VQA-Caption models ($z_I$ and $z_C$ in Section~\ref{sec:vqa_model}) as image and caption encodings in place of the VQA activations ($u_I$ and $u_C$ in Section~\ref{sec:vqa_grounded_representations}). 
Using these hidden activations of the VQA models is conceptually similar to using the hidden activations of CNNs pretrained on ImageNet as features~\cite{Donahue_2013}.
These features achieve 46.8\% caption retrieval recall@1 and 35.2\% image retrieval recall@1 for score-level fusion, and 49.3\% caption retrieval recall@1 and 37.9\% image retrieval recall@1 for representation-level fusion which are as good as our semantic features $u_I$ and $u_C$.
%It shows that we can leverage knowledge in VQA corpora in a non-semantic way using our approaches for image-caption ranking.
This shows that our semantically meaningful features, $u_I$ and $u_C$, performs as well as their corresponding non-sematic representations $z_I$ and $z_C$ using both score-level fusion and representation-level fusion. Note that such hidden activations may not always be available in different VQA models and the semantic features have the added benefit of being interpretable (\eg, Figure~\ref{fig:qual_vqa}).

\subsection{Ablation study}
\label{sec:results_ablation}

As an ablation study, we compare the following four models: 1) full representation-level fusion: our full N = 3000 representation-level fusion model that includes both image and caption VQA representations; 2) caption-only representation-level fusion: the same representation-level fusion model but using the VQA representation only for the caption, \cameraready{$v_C$}, and not for the image; 3) image-only representation-level fusion: the same model but using the VQA representation only for the image, \cameraready{$v_I$}, and not for the caption; 4) deeper VQA-agnostic: The N = 0 representation-level fusion model described earlier that does not use VQA representations for \xiao{neither} the image nor the caption.

%\ablation{We compare the performance of using only representation of the caption $t_C$ (caption-only) or the image $t_I$ (image-only) as opposed to using both in our full $N=3000$ representation-level fusion model (full representation-level fusion) and using neither in the $N=0$ deeper VQA-agnostic baseline (deeper VQA-agnostic). }

Table~\ref{table:ablation} summarizes the results. We see that incrementally adding more VQA-knowledge improves performance. Both caption-only and image-only models outperform the $N=0$ deeper VQA-agnostic baseline. The full representation-level fusion model which combines both representations yields the best performance.

\subsection{The role of VQA and caption annotations}
\label{sec:results_caption}

\begin{table}
\caption{Ablation study evaluating the gain in performance as more VQA-knowledge is incorporated in the model}
\centering
\begin{tabular}{ l |c c c|c c c }
\toprule
\multicolumn{7}{c}{\textbf{MSCOCO}}\\
\midrule
\multicolumn{1}{c}{Approach}&  \multicolumn{3}{c}{Caption Retrieval} & \multicolumn{3}{c}{Image Retrieval}\\
    & R@1   & R@5    & R@10   & R@1 & R@5 & R@10 \\
\midrule
Deeper VQA-agnostic  &  45.8 & 76.8  & 86.1  & 33.6 & 67.8 & 81.0\\

Caption-only representation-level fusion  &  47.3 & 77.3  & 86.6  & 35.5 & 69.3 & 81.9\\

Image-only representation-level fusion  &  47.0 & 80.0  & 89.6  & 36.4 & 70.1 & 82.3\\

Full representation-level fusion  &  \textbf{50.5}  &  \textbf{80.1}   & \textbf{89.7}  & \textbf{37.0} & \textbf{70.9} & \textbf{82.9}\\
					
\bottomrule
\end{tabular}
\label{table:ablation}
\end{table}

In this work we transfer knowledge from one vision-language task (\ie VQA) to another (\ie image-caption ranking). However, VQA annotations and caption annotations serve different purposes.

The target language to be retrieved is caption language, and not VQA language. ~\cite{Antol_2015} showed qualitatively and quantitatively that the two languages are statistically quite different (in terms of information contained, and in terms of nouns, adjectives, verbs, etc. used). As a result, VQA can not be thought of as providing additional ``annotations'' for the captioning task. Instead, VQA provides different perspectives/views of the images (and captions). It provides an additional feature representation. To better utilize this representation for an image-caption ranking task, one would still require sufficient ground truth caption annotations for images. In fact, with varying amounts of ground truth (caption) annotations, the VQA-aware representations show improvements in performance across the board. See Figure~\ref{fig:caption_scaling} (left).

A better analogy of our VQA representation is hidden activations (\eg, fc7) from a CNN trained on ImageNet. Having additional ImageNet annotations would improve the fc7 feature. But to map this fc7 feature to captions, one would still require sufficient caption annotations. Conceptually, caption annotations and category labels in ImageNet play two different roles. The former provides ground truth for the target task at hand (image-caption ranking), and having additional annotations for the target application typically helps. The latter helps learn a better image representation (which may provide improvements in a variety of tasks). 
%\cameraready{Evaluating if the VQA representation can be transferred to other tasks beyond image-caption ranking is part of future work.}

\subsection{Number of question-answer pairs}
\label{sec:results_qa_pairs}
\begin{figure}
\begin{center}
   \includegraphics[width=0.95\linewidth]{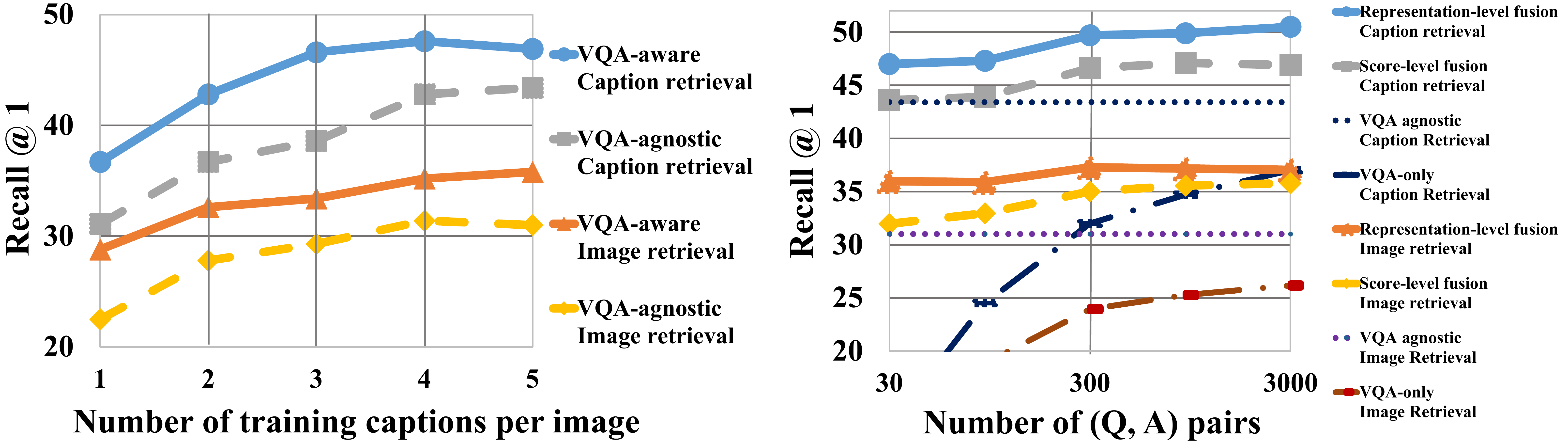}
\end{center}
   \caption{\textbf{Left}: \cameraready{caption retrieval and image retrieval performances of the VQA-agnostic model compared with our $N=3000$ score-level fusion VQA-aware model trained using 1 to 5 captions per image.} The VQA representations in the VQA-aware model provide consistent performance gains. \textbf{Right}: caption retrieval and image retrieval performances of our score-level fusion and representation-level fusion approaches with varying number of $(Q,A)$ pairs used for feature extraction.}
\label{fig:caption_scaling}
\end{figure}

Our VQA-grounded representations extract image and caption features based on question-answer pairs. It is important for there to be enough question-answer pairs to cover necessary aspects for image-caption ranking. We experiment with using $N=30, 90, 300, 900,$ $3000$ $(Q,A)$ pairs (or facts) for both score-level and representation-level fusion. Figure~\ref{fig:caption_scaling} (right) shows caption and image retrieval performances of our approaches with varying $N$. Performance \cameraready{of both score-level and representation-level fusion approaches} improve quickly from $N=30$ to $N=300$, and then starts to level off after $N=300$.

An alternative to sampling 3 question-answer pairs per image on 1,000 images to get $N=3000$ questions is to sample 1 question-answer pair per image from 3,000 images. 
Sampling multiple $(Q,A)$ pairs from the same image provides correlated $(Q,A)$ pairs. For example ($Q$: What are these animals? $A$: Giraffes) and ($Q$: Would this animal fit in a house? $A$: No). Using such correlated $(Q,A)$ pairs, the model could potentially better predict if there is a giraffe in the image by jointly reasoning if the animal looks like a giraffe and the if the animal would fit in a house, if the VQA and VQA-Caption models have not already picked up such correlations.
In experiments, sampling 3 question-answer pairs per image for correlated $(Q,A)$ pairs does not significantly outperform sampling 1 question-answer pair per image \cameraready{which performs (47.7\%, 35.4\%) (image, caption) recall@1 using $N=3000$ score-level fusion}, so we hypothesize that our VQA and Caption-QA models have already captured such correlations. 

%Exploring active learning and \xiao{other} human-in-the-loop applications is part of future work.

\section{Conclusion}
\label{sec:discussion}

%We evaluated our score-level fusion and representation-level fusion VQA-aware models for image-caption ranking. Our approach significantly outperforms the VQA-agnostic model and set new state-of-the-arts on MSCOCO by large margins. 
%The VQA dataset~\cite{Antol_2015} provides rich multi-modal knowledge for image-caption ranking that are complementary to image-caption ranking datasets.

%Assessing possiblities of VQA question-answer pairs provides rich interpretations for images and captions, and opens up research directions for search~\cite{Kovashka_2012}, zero-shot learning~\cite{Lampert_2009}, human-in-the-loop~\cite{Branson_2010} \etc.

%Improving VQA and VQA-Caption models, end-to-end learning, an attention mechanism for image-specific question-answer pair selection/generation might further improve the performance of our model.

VQA corpora provide rich multi-modal information that is complementary to knowledge stored in image captioning corpora. In this work we take the novel perspective of viewing VQA as a ``feature extraction'' module that captures VQA knowledge. We propose two approaches -- score-level and representation-level fusion -- to integrate this knowledge into an existing image-caption ranking model. We set new state-of-the-art by improving caption retrieval by 7.1\% and image retrieval by 4.4\% on MSCOCO.

%Each feature dimension in our VQA-grounded activation is interpretable -- the plausibility of the corresponding fact (question-answer pair) being true for an image or caption. \xiao{That opens opportunity to better understand the model. Exploring active learning and human-in-the-loop applications is part of future work. }

%\cameraready{Modeling whether a question applies to the target images and captions as studied in contemporary work~\cite{Ray_2016} reveals the premise of images and captions and may be informative to image-caption ranking. It can be incorporated as an additional feature in our approach.} 
Improved individual modules, \ie, VQA models and VQA-agnostic image-caption ranking models, end-to-end training, and an attention mechanism that selects question-answer pairs (facts) in an image-specific manner may further improve the performance of our approach. 

\section{Acknowledgment}

%We thank Stanislaw Antol and Ramakrishna Vedantam for discussions. We thank Jamie Ryan Kiros for help with the VQA-agnostic baseline code. 

This work was supported in part by the Allen Distinguished Investigator awards by the Paul G. Allen Family Foundation, a Google Faculty Research Award, a Junior Faculty award by the Institute for Critical Technology and Applied Science (ICTAS) at Virginia Tech, a National Science Foundation CAREER award, an Army Research Office YIP award, and Office of Naval Research YIP award to D. P. The views and conclusions contained herein are those of the authors and should not be interpreted as necessarily representing the official policies or endorsements, either expressed or implied, of the U.S. Government or any sponsor.

\appendix
\section*{Appendix}

\section{VQA Models}

\begin{figure}
\begin{center}
   \includegraphics[width=1\linewidth]{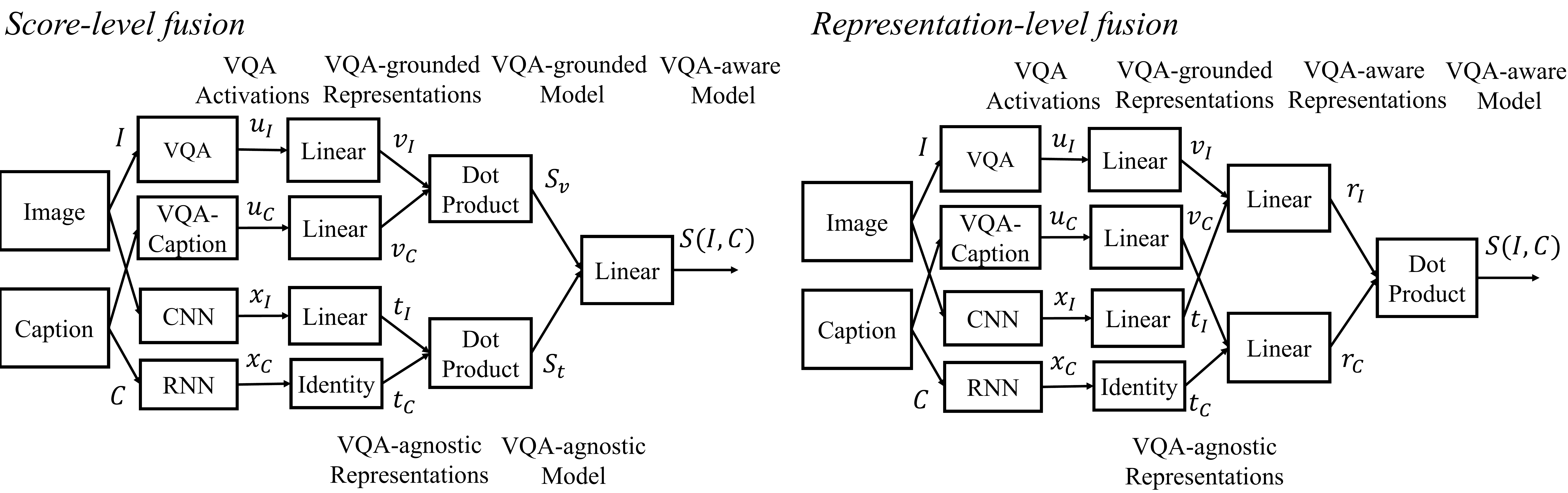}
\end{center}
   \caption{Our VQA and VQA-Caption network architectures. Details of the VQA and VQA-Caption models can be found in our paper. }
\label{fig:vqa}
\vspace{-10pt}
\end{figure}

Figure~\ref{fig:vqa} illustrates the network architectures of our VQA and VQA-Caption models. 

\section{Results on MSCOCO--5K, Flickr8k and Flickr30k}

\xiao{Table~\ref{table:mscoco5k} shows results on MSCOCO using all 5,000 test images following the protocol of~\cite{Karpathy_2015}. Retrieving from 5,000 test images is more challenging than retrieving from 1,000 test images so the performances of all models are lower. However, the trends are consistent with results on 1,000 test images reported in the main paper. Our score-level fusion model achieves \xiao{22.8\%} caption retrieval R@1 and \xiao{15.5\%} image retrieval R@1, outperforming the VQA-agnostic model by \xiao{4.7\%} and \xiao{2.8\%}. Our representation-level fusion model achieves \xiao{23.5\%} caption retrieval R@1 and \xiao{16.7\%} image retrieval R@1.}

Flickr8k~\cite{Hodosh_2013} and Flickr30k~\cite{Young_2014} consist of 8,000 and 30,000 images, respectively, collected from Flickr. Each image in Flickr8k and Flickr30k is annotated with 5 image captions. Following the evaluation protocol of~\cite{Karpathy_2015} we use 1,000 images for validation, 1,000 images for testing, the rest for training and report recall@(1, 5, 10) for caption retrieval and image retrieval on test.

Table~\ref{table:flickr8k} and Table~\ref{table:flickr30k} show results on Flickr8k and Flickr30k dataset, respectively. Our VQA-aware model shows consistent improvements over the VQA-agnostic model on both datasets. On Flickr8k our score-level fusion approach achieves \xiao{24.3\%} caption retrieval R@1 and \xiao{17.2\%} image retrieval R@1, which outperforms the VQA-agnostic model by \xiao{2.0\%} and \xiao{2.3\%}. On Flickr30k our score-level fusion approach achieves \xiao{33.9\%} caption retrieval R@1 and \xiao{24.9\%} image retrieval R@1, which outperforms the VQA-agnostic model by \xiao{4.1\%} and \xiao{2.9\%}. 

Note that the VQA and VQA-Caption models are trained on MSCOCO which is a different dataset. Yet, they consistently improve image-caption ranking on Flickr8k and Flickr30k. It shows that our VQA-grounded image and caption representations generalize across datasets. Fine-tuning on these datasets, and incorporating our approach on top of state-of-the-art captioning approaches on these datasets (Instead of~\cite{Kiros_2015} which is state-of-the-art on MSCOCO but not Flickr) may further improve our performance.

Both Flickr8k and Flickr30k are smaller compared with the MSCOCO dataset. Our representation-level fusion model overfits to the training sets despite using dropout.

\section{Qualitative examples}

\begin{figure}
\begin{center}
   \includegraphics[width=0.95\linewidth]{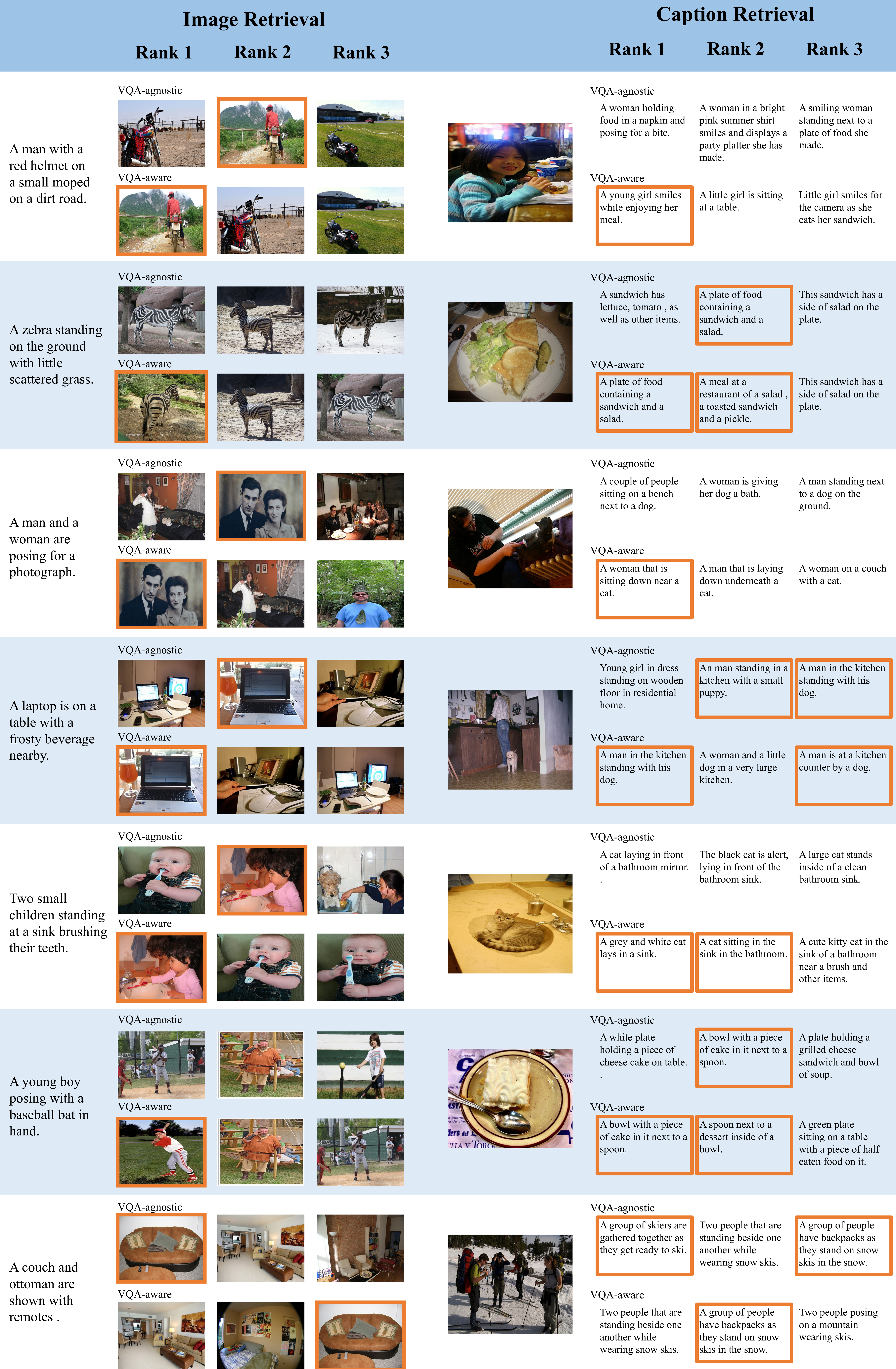}
\end{center}
   \caption{Qualitative results of image retrieval and caption retrieval at rank 1, 2 and 3 using our $N=3,000$ score-level fusion VQA-aware model and the baseline VQA-agnostic model. The true target images and captions are highlighted.}
\label{fig:qual_add}
\end{figure}

Fig.~\ref{fig:qual_add} shows additional qualitative examples of image retrieval and caption retrieval using our $N=3,000$ score-level fusion model (VQA-aware) and the baseline VQA-agnostic model (VQA-agnostic).

\begin{table*}[!htbp]
\caption{Results on MSCOCO using all 5,000 test images}
\centering
\begin{tabular}{ l |c c c|c c c }
\toprule
\multicolumn{7}{c}{\textbf{MSCOCO 5K test images}}\\
\midrule
\multicolumn{1}{c}{Approach}&  \multicolumn{3}{c}{Caption Retrieval} & \multicolumn{3}{c}{Image Retrieval}\\
    & R@1   & R@5    & R@10   & R@1 & R@5 & R@10 \\
\midrule
Random  &  0.1 & 0.5  & 1.0  & 0.1 & 0.5 & 1.0 \\
DVSA~\cite{Karpathy_2015}  &  16.5 & 39.2  & 52.0  & 10.7 & 29.6 & 42.2 \\
FV (GMM+HGLMM)~\cite{Klein_2015} & 17.3 & 39.0 & 50.2 & 10.8 & 28.3 & 40.1 \\
Kiros \etal~\cite{Kiros_2015} (VQA-agnostic)   &  18.1 & 43.5  & 56.8  & 12.7 & 34.0 & 47.3\\
\midrule
N=3000 score-level fusion VQA-grounded only  & 15.7 & 37.9 & 50.3 & 11.0 & 29.5 & 42.0 \\
N=3000 score-level fusion VQA-aware &  22.8  &  49.8 & 63.0  & 15.5 & 39.1 & 52.6 \\
\midrule
N=0 representation-level fusion VQA-agnostic  &  20.6 & 47.1  & 60.3  & 14.9 & 37.8 & 50.9\\
N=3000 representation-level fusion VQA-aware  &  \textbf{23.5}  &  \textbf{50.7}   & \textbf{63.6}  & \textbf{16.7} & \textbf{40.5} & \textbf{53.8}\\
					
\bottomrule
\end{tabular}
\label{table:mscoco5k}
\end{table*}

\begin{table*}[!htbp]
\caption{Results on Flickr8k dataset}
\centering
\begin{tabular}{ l |c c c|c c c }
\toprule
\multicolumn{7}{c}{\textbf{Flickr8k}}\\
\midrule
\multicolumn{1}{c}{Approach}&  \multicolumn{3}{c}{Caption Retrieval} & \multicolumn{3}{c}{Image Retrieval}\\
    & R@1   & R@5    & R@10   & R@1 & R@5 & R@10 \\
\midrule
Random  &  0.1 & 0.5  & 1.0  & 0.1 & 0.5 & 1.0 \\
DVSA~\cite{Karpathy_2015}  &  16.5 & 40.6  & 54.2  & 11.8 & 32.1 & 43.8 \\
FV (GMM+HGLMM)~\cite{Klein_2015} & \textbf{31.0} & \textbf{59.3} & \textbf{73.7} & \textbf{21.3} & \textbf{50.0} & \textbf{64.8} \\
$m$-RNN-AlexNet~\cite{Mao_2015}   &  14.5 & 37.2  & 48.5  & 11.5 & 31.0 & 42.4 \\
$m$-$CNN_{ENS}$~\cite{Ma_2015}   &  24.8 & 53.7   & 67.1  & 20.3 & 47.6 & 61.7\\
Kiros \etal~\cite{Kiros_2015} (VQA-agnostic)   &  22.3 & 48.7  & 59.8  & 14.9 & 38.3 & 51.6\\
\midrule
N=3000 score-level fusion VQA-grounded only  & 10.5 &  31.5 & 42.7 &  7.6 & 22.8   & 33.5 \\
N=3000 score-level fusion VQA-aware &  24.3  &  52.2   & 65.2  & 17.2 & 42.8 & 57.2 \\
\bottomrule
\end{tabular}
\label{table:flickr8k}
\end{table*}

\begin{table*}[!htbp]
\caption{Results on Flickr30k dataset}
\centering
\begin{tabular}{ l |c c c|c c c }
\toprule
\multicolumn{7}{c}{\textbf{Flickr30k}}\\
\midrule
\multicolumn{1}{c}{Approach}&  \multicolumn{3}{c}{Caption Retrieval} & \multicolumn{3}{c}{Image Retrieval}\\
    & R@1   & R@5    & R@10   & R@1 & R@5 & R@10 \\
\midrule
Random  &  0.1 & 0.5  & 1.0  & 0.1 & 0.5 & 1.0 \\
DVSA~\cite{Karpathy_2015}  &  22.2 & 48.2  & 61.4  & 15.2 & 37.7 & 50.5 \\
FV (GMM+HGLMM)~\cite{Klein_2015} & 35.0 & 62.0 & 73.8 & 25.0 & 52.7 & 66.0 \\
RTP (weighted distance)~\cite{Plummer_2015} & \textbf{37.4} & 63.1 & 74.3 & 26.0 & 56.0 & 69.3 \\
$m$-RNN-vgg~\cite{Mao_2015}   &  35.4 & 63.8  & 73.7  & 22.8 & 50.7 & 63.1 \\
$m$-$CNN_{ENS}$~\cite{Ma_2015}   &  33.6 & \textbf{64.1} & \textbf{74.9}  & \textbf{26.2} & \textbf{56.3} & \textbf{69.6} \\
Kiros \etal~\cite{Kiros_2015} (VQA-agnostic)   &  29.8 & 58.4  & 70.5  & 22.0 & 47.9 & 59.3 \\
\midrule
N=3000 score-level fusion VQA-grounded only  & 17.6 &  40.5 & 51.2 &  12.7 & 31.9   & 42.5 \\
N=3000 score-level fusion VQA-aware &  33.9  &  62.5   & 74.5  & 24.9 & 52.6 & 64.8 \\
\bottomrule
\end{tabular}
\label{table:flickr30k}
\end{table*}

\clearpage
\bibliographystyle{splncs03}
\bibliography{main}
\end{document}